\documentclass[letterpaper, 10 pt, conference]{ieeeconf}
\usepackage{cite}
\usepackage{amsmath,amssymb,amsfonts}
\usepackage[ruled,linesnumbered]{algorithm2e}
\usepackage{algorithmic}
\usepackage{graphicx}
\usepackage{subfigure}
\usepackage{graphics}
\usepackage{textcomp}
\usepackage{xcolor}
\usepackage{mathptmx} 
\usepackage{times} 
\usepackage{bm}
\usepackage{multirow}
\usepackage{booktabs}
\usepackage{array}
\usepackage{tabularx,booktabs}
\usepackage{overpic}
\usepackage{url}
\usepackage[bookmarks=false]{hyperref}
\usepackage{rotating}
\usepackage{mathtools}
\usepackage{cite}
\usepackage{multirow}
\IEEEoverridecommandlockouts                              

\usepackage{verbatim}

\title{\LARGE \bf 
        Crowd-Aware Robot Navigation for Pedestrians with Multiple Collision Avoidance Strategies via Map-based Deep Reinforcement Learning
}

\author{Shunyi Yao$^{1, *}$, Guangda Chen$^{1, *}$, Quecheng Qiu$^2$, Jun Ma$^2$,  Xiaoping Chen$^1$, and Jianmin Ji$^{1, \dag}$ 
\thanks{The work is partially supported by the National Key Research and Development Program of China (No. 2018AAA0100500), CAAI-Huawei MindSpore Open Fund, Anhui Provincial Development and Reform Commission 2020 New Energy Vehicle Industry Innovation Development Project ``Key System Research and Vehicle Development for Mass Production Oriented Highly Autonomous Driving'', and Key-Area Research and Development Program of Guangdong Province 2020B0909050001.
}
\thanks{$^1$ School of Computer Science and Technology, University of Science and Technology of China (USTC), Hefei 230026, China}
\thanks{$^2$ School of Data Science, USTC, Hefei 230026, China}
\thanks{ $^*$ These authors contributed equally to the work. {\tt\small \{ustcysy, cgdsss\}@mail.ustc.edu.cn}}
\thanks{ $^\dag$ Corresponding author. {\tt\small jianmin@ustc.edu.cn}}}
\begin{document}

\maketitle
\thispagestyle{empty}
\pagestyle{empty}

\begin{abstract}
It is challenging for a mobile robot to navigate through human crowds. 
Existing approaches usually assume that pedestrians follow a predefined collision avoidance strategy, like social force model (SFM) or optimal reciprocal collision avoidance (ORCA). However, their performances commonly need to be further improved for practical applications, where pedestrians follow multiple different collision avoidance strategies.
In this paper, we propose a map-based deep reinforcement learning approach for crowd-aware robot navigation with various pedestrians.
We use the sensor map to represent the environmental information around the robot,  including its shape and observable appearances of obstacles.
We also introduce the pedestrian map that specifies the movements of pedestrians around the robot.
By applying both maps as inputs of the neural network, we show that a navigation policy can be trained to better interact with pedestrians following different collision avoidance strategies. 
We evaluate our approach under multiple scenarios both in the simulator and on an actual robot. 
The results show that our approach allows the robot to successfully interact with various pedestrians and outperforms compared methods in terms of the success rate.
\end{abstract}


\section{Introduction}
\label{sec:intro}

Many practical applications of mobile robots, like domestic service robots~\cite{forlizzi2006service} and office delivery robots~\cite{simmons1997layered}, require them to navigate through human crowds.
It is a tough challenge, as pedestrians around the robot can not be simply considered as regular static or dynamic obstacles and the robot needs to interact with a crowd of people in a socially compliant manner. 
Existing crowd navigation approaches commonly make assumptions on the movements~\cite{van2008reciprocal} or the collision avoidance strategies~\cite{helbing1995social,van2011reciprocal} of pedestrians. 
However, the performance of these approaches is usually limited when pedestrians behave differently. 

On the other hand, deep reinforcement learning (DRL) approaches have been applied for robot navigation with promising results~\cite{tai2017virtual,xiao2020motion,chen12020robot}.
Recently, DRL approaches are also applied for crowd navigation. 
Everett et~al.~\cite{everett2018motion} and Chen et~al.~\cite{chen2019crowd} assume that pedestrians follow a certain collision avoidance strategy, i.e., optimal reciprocal collision avoidance (ORCA)~\cite{van2011reciprocal}. 
Moreover, both approaches require perfect sensing, as the inputs of their networks include exact positions and movements of surrounding obstacles.
Guldenring et~al.~\cite{guldenringlearning} assume pedestrians following social force model (SFM)~\cite{helbing1995social} and use the sensor data directly.
Notice that, above approaches do not distinguish the information between regular obstacles and pedestrians, which increases the difficulty for the trained model to handle pedestrians with multiple collision avoidance strategies. 

In this paper, we propose a map-based DRL approach for crowd navigation with pedestrians following multiple different collision avoidance strategies.
We find out that it is more effective for the trained model to handle various pedestrians by distinguishing the information between regular obstacles and pedestrians.
In particular, following our previous work~\cite{chen2020distributed} for robot navigation among regular obstacles, we use the sensor map to represent the environmental information around the robot, including its shape and observable appearances of obstacles, which can be directly generated from robot's sensor information.
We also introduce the pedestrian map that specifies the movements of pedestrians around the robot in the local grid map. 
Then we apply the proximal policy optimization (PPO) algorithm~\cite{schulman2017proximal} to train a convolutional neural network that directly maps both maps and the robot's target pose into low-level robot control commands.
Note that, the information of pedestrians is specified by the pedestrian map, which can ease the training of the network to interact with various pedestrians.
We first train the neural network in a specified simulator with static obstacles and pedestrians following two strategies, i.e., SFM and ORCA, and then we deploy the trained model to an actual robot for its navigation.

We evaluate our approach under multiple scenarios both in the simulator and on an actual robot. 
Experimental results show that, by applying both the sensor map and the pedestrian map, the network allows the robot not only to avoid regular obstacles but also pedestrians with different collision avoidance strategies. 
We compare our approach to a dynamic obstacle avoidance method with perfect sensing and a DRL method only using the sensor map.
The results show that our approach outperforms these methods in terms of the success rate.
We deploy the trained model to a robot and evaluate its performance in real-world environments, like corridors and halls.
The demonstration video is also available.
Our main contributions are summarized as follows:
\begin{itemize}
        \item We propose a DRL-based crowd navigation method that allows pedestrians to follow multiple different collision avoidance strategies.
		The experimental results show that the approach is effective and outperforms
		compared methods in terms of the success rate. 
		\item We apply both the sensor map and the pedestrian map as the inputs of the neural network. 
		We show that, by distinguishing the information between regular obstacles and pedestrians, the network allows the robot not only to avoid regular obstacles but also various pedestrians.
		\item We show that, the map-based approach allows the robot to generate the sensor map and the pedestrian map from multiple sensor data or sensor fusion results~\cite{chen2020distributed}, which make it to be effective and easy to be deployed to an actual robot. 
\end{itemize}

\section{Related Work}
\label{rw}


Existing dense crowd navigation methods can be generally divided into two categories, i.e., trajectory-based and rule-based.
A trajectory-based method first predicts the future path of other agents in the scene,
and then selects the optimal path based on the prediction~\cite{chen2018robot,luders2011probabilistically}. 
This type of methods normally require a high computational cost and their performances are significantly affected by the accuracy of the prediction.

On the other hand, a rule-based method commonly assumes that all agents in the environment follow the same collision avoidance strategy.
Helbing and Molnar~\cite{helbing1995social} propose social force model (SFM), which assumes that pedestrians are subject to three main forces in the environment:
the influence of the driving force, the force between people, and the force between people and obstacles.
The resultant force of these forces produces the acceleration of the pedestrian's movement.
Reciprocal velocity obstacles (RVO)~\cite{van2008reciprocal} and ORCA~\cite{van2011reciprocal} assume that the agents can perfectly perceive the movements of others and cooperate with each other to generate collision-free movements respectively.
Note that, the performance of these methods is usually limited when pedestrians behave differently.

Recently, imitation learning methods have been applied for crowd navigation.
Kretzschmar et al. \cite{kretzschmar2016socially} propose an approach that allows a mobile robot to learn the behavior of interacting agents such as pedestrians from demonstration via Inverse Reinforcement Learning.
Liu et al. \cite{liu2018map} imitate the well-tuned planner, which considers the costmap\footnote{\url{http://wiki.ros.org/costmap_2d}.} of the robot as the input and outputs robot's control commands, i.e., linear and angular velocities.
Tai et al. \cite{tail2018socially} train the neural network in Gazebo simulator\footnote{\url{http://gazebosim.org/}.} with simulation pedestrians driven by SFM. 
However, these methods require a large number of labeled data for training.

DRL methods are also applied for crowd navigation. 
Fan et al. \cite{fan2020distributed} consider the sensor data of the 2D laser scanner as the input of the policy network and trains the network in simulation environments with multiple robots driven by the same policy network, which allows the robots to properly interact with each other. 
Similarly, Chen et al. \cite{chen2017socially} consider the movement information of other agents as the input of the policy network and trains the network with multiple robots driven by the same network in~\cite{chen2017decentralized}.
In our previous work \cite{chen2020distributed}, the egocentric sensor map is considered as the input of the policy network, which is also trained with multiple robots driven by the same network.
Notice that, these methods do not distinguish the information between regular obstacles and pedestrians.

On the other hand, G{\"u}ldenring et al.~\cite{guldenringlearning} use SFM to drive pedestrians in their training environments.
Meanwhile, the shapes and movements of pedestrians' legs are carefully modeled in the simulator to distinguish pedestrians with other obstacles. 
Chen et al.~\cite{chen2019crowd} use ORCA to drive pedestrians in their training environments and considers the movement data of pedestrians as the input, where regular obstacles are considered as special agents that do not move. 
Note that, above methods assume that pedestrians only follow a certain collision avoidance strategy, which is often not appropriate in practice. 

Liu et al. \cite{liu2020robot} use two strategies, i.e., ORCA and moving forward, to drive pedestrians respectively. However, it is hard to generate the approach for pedestrians with other collision avoidance strategies, like SFM.
In this paper, we distinguish the information between regular obstacles and pedestrians and propose a map-based DRL approach for crowd navigation with pedestrians following multiple different strategies.


\section{Preliminaries}

In DRL, the problem is specified as a Markov decision process (MDP). 
An MDP is a tuple $M=(S, A, P, R, \gamma)$, 
where $S$ is the state space,
$A$  is the action space,
$P$ represents the transition probability between states,
$R$ is the reward function, and
$\gamma$ is the discount factor in $(0,1)$.
The purpose of DRL is to find a strategy $\pi^{*}$ that maximizes the expected cumulative return of each trajectory $\tau$.

We use proximal policy optimization (PPO)~\cite{schulman2017proximal} to train our crowd navigation policy.
PPO maintains two networks during the training, i.e., a policy network and a value network. 
It uses importance sampling when updating the parameters, so that each sampled data can be applied multiple times. 
PPO also introduces the clip operation when the policy is updated to reduce the gap between the behavior policy and the target policy.
We use generalized advantage estimator (GAE)~\cite{schulman2015high} to estimate the advantage function used in PPO.

\section{Approach}
\label{approach}

In this section, we first specify key components of DRL for crowd navigation.
Then, we describe its training process.

\subsection{Reinforcement Learning Components}

\subsubsection{Observation space}

An observation is composed of three parts, i.e., the sensor map, the three-channel pedestrian map, and the robot's target pose.

\textit{Sensor map} is specified by an egocentric local grid map $\textbf{M}_{sens}$ of the robot. As illustrated in Fig.~\ref{network}, the sensor map represents the environmental information around the robot, including its shape and observable appearances of obstacles.
The sensor map is constructed from a costmap that can be generated from various sensor data, like outputs of 2D lasers or depth cameras~\cite{chen2020distributed}. 

\textit{Pedestrian map} is a local grid map $\textbf{M}_{ped}$ with three channels that indicates the location and speed of pedestrians around the robot.
As illustrated in Fig.~\ref{network}, for a pedestrian map, the first channel specifies locations of surrounding pedestrians and the rest two channels specify the speeds of corresponding pedestrians for $x$-axis and $y$-axis, respectively.

\textit{Target pose} $g^{t}=(\,x^t,\,y^t,\,\alpha^t\,)$ at time step~$t$ consists of the target position $(x^t,\, y^t)$ and the target orientation $\alpha^t$ of the robot.

\subsubsection{Action space}

In this paper, we implement the approach on a differential drive robot that follows desired speed commands. 
Then an action $a^t$ at time step~$t$ consists of a linear velocity $v^t$ and an angular velocity $\omega^t$, i.e., $a^t = ( v^t,\, \omega^t)$.
We implement both discrete and continuous actions for the robot. 
In specific, for discrete actions, we set a linear velocity $v^t\in \{ 0.0, 0.2, 0.4, 0.6\}$ and an angular velocity $\omega^t \in \{0.9,-0.6,-0.3,0.0,0.3,0.6,0.9\}$.
For continuous actions, we set $v^t\in [0,\, 0.6]$ and $\omega^t \in [-0.9,\, 0.9]$. 
Both discrete and continuous actions can be directly performed by the differential robot in our experiments.
Notice that, $v^t \geq 0$, i.e., moving backwards is not allowed, due to the lack of rear sensors.

\subsubsection{Reward function}

In crowd navigation, the objective of a navigation policy is to minimize the arriving time of the robot without collision. 
In our approach, we specify the reward $r^t$ at time step~$t$ as
\begin{align*}
r^{t} &=r_{goal}^{t}+r_{safe}^{t}+r_{step}^{t}+r_{shaping}^{t},\\
r_{goal}^{t} &=\left\{\begin{array}{ll}
        r_{arr} & \text {if target is reached,} \\
        0 & \text { otherwise,}
        \end{array}\right.\\
r_{safe}^{t} &=\left\{\begin{array}{ll}
        r_{col} & \text { if collision, } \\
        -\varepsilon_{1}\left(1-d_{\min }^{t}\right) & \text { if } d_{min }^{t}<1, \\
        0 & \text { otherwise,}
        \end{array}\right.   \\
r_{shaping}^t &=\varepsilon_{2}\left(\left\|\mathbf{p}^{t-1}-\mathbf{p}_{g}\right\|-\left\|\mathbf{p}^{t}-\mathbf{p}_{g}\right\|\right),
\end{align*}
where $r_{arr}>0$ and $r_{goal}^t$ specifies the reward when the robot arrives its target, $r_{col}<0$ denotes the penalty for the collision, $-\varepsilon_{1}\left(1-d_{\min }^{t}\right)$ denotes the penalty when the robot is close to a pedestrian, i.e., $d_{min}^t$ denotes the minimal distance from the robot to its closest pedestrian, $\varepsilon_{1}$ is a hyper-parameter, $r_{step}^{t} <0$ denotes a small penalty to encourage short paths, $\textbf{p}^t$, $\textbf{p}_g$ denote the positions of the robots and its target, $\varepsilon_{2}$ is a hyper-parameter, and the reward shaping item $r_{shaping}^t$ encourages the robot to move toward the target.

In our experiments, we set $r_{arr}= 500$, $r_{col}=-500$, $\varepsilon_{1}=50$, $\varepsilon_{2}=200$, and $r^t_{step}=-5$.

\subsubsection{Network architecture}

\begin{figure}
        \centering
        \begin{overpic}[width=\linewidth]{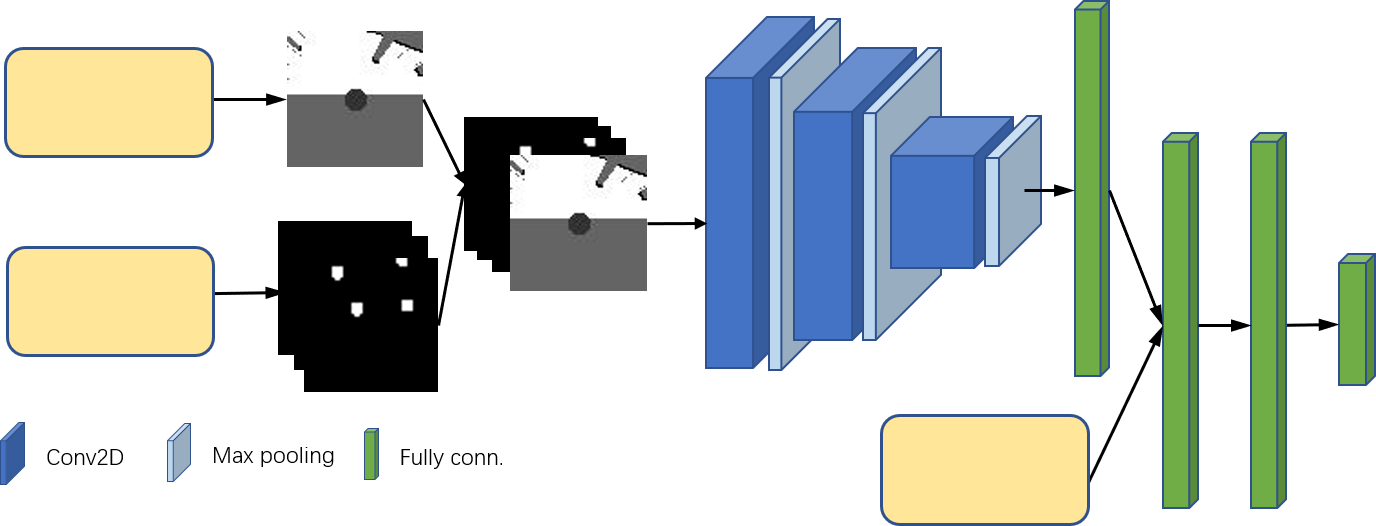}
        \put(1.6,31.4){\scriptsize Sensor}
         \put(1.6,28.4){\scriptsize Map}
         \put(1.6,17){\scriptsize Pedestrian}
         \put(1.6,14){\scriptsize Map}
         \put(68,4.2){\scriptsize Target}
         \put(68,1.2){\scriptsize Pose}
        \end{overpic}
        \caption{The architecture of the crowd navigation policy network.}
        \label{network}
\end{figure}
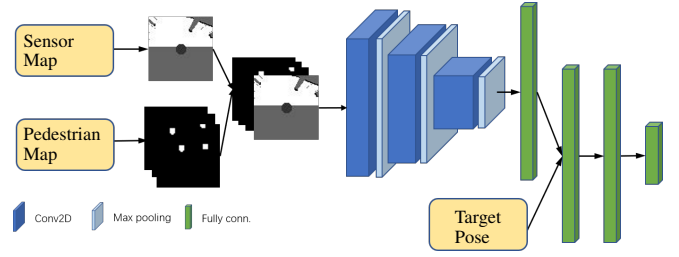

The architecture of the convolutional network for the crowd navigation policy in PPO is shown in Fig.~\ref{network}. 
The input of the network consists of the sensor map, the pedestrian map, and the target pose. 
For discrete actions, the network outputs a 28-dimensional vector from a softmax layer to choose the pair of linear and angular velocities.
For continuous actions, the network outputs the mean of the action sampled from a Gaussian distribution.

In particular, the network first produces feature maps for the corresponding sensor map and the pedestrian map using three convolutional layers and three max pooling layers.
Followed by a fully-connected layer with 512 units, these feature maps are converted to a 512 dimensional vector.
The network also projects the local goal to a 3 dimensional vector.
Then the network combines both vectors and feeds them to two fully-connected layers with 512 units. 
At last, for discrete actions, the network uses a softmax layer with 28 units to choose the pair of linear and angular velocities from corresponding values.
For continuous actions, the network applies a fully-connect layer with 2 units without activation to produce the mean of the linear velocity $v^t$ and the mean of the angular velocity $\omega^t$.
Then the continuous actions is sampled from the Gaussian distribution $\mathcal{N}(a_{mean}^t,\,a_{logstd}^t)$, where $a_{logstd}^t $ is  the  log standard deviation generated by a standalone network and $a_{mean}^t = (v^t, \omega^t)$.
A clip function is also applied to ensure that the resulting actions are valid in the action space.

The value network has the same architecture as the policy network, except its last layer is modified to only output the value of the state.

\subsection{Training with Multiple Strategies and Environments}\label{sec:train}

We train the network in environments that are constructed by a customized simulator based on OpenCV\footnote{\url{https://opencv.org/}.}. 
We extend the simulator from the one used in \cite{chen2020distributed} to add pedestrians driven by different collision avoidance strategies.
In particular, we add the shapes of pedestrians' legs and corresponding walking movements as in \cite{guldenringlearning} to the simulator. 
We also implement two collision avoidance strategies, i.e., ORCA\footnote{\url{https://github.com/snape/RVO2}.} and SFM\footnote{\url{https://github.com/chgloor/pedsim}.} to drive pedestrians in the simulator. 

\begin{figure}
        \centering
        \subfigure[Random scenario]{\includegraphics[width = 0.48\linewidth]{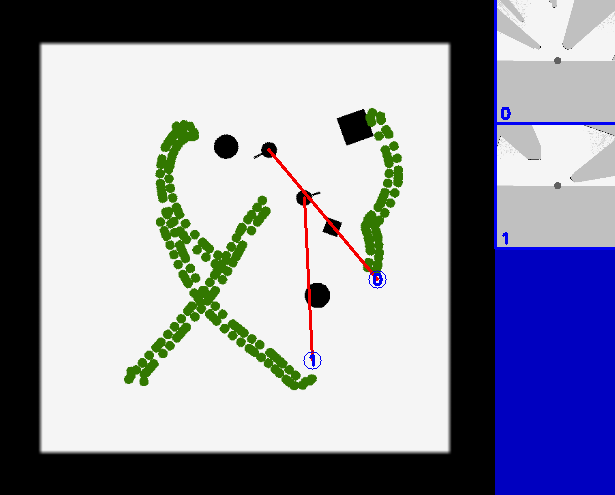}\label{fig:example:e1}}
        \subfigure[Circular scenario]{\includegraphics[width = 0.48\linewidth]{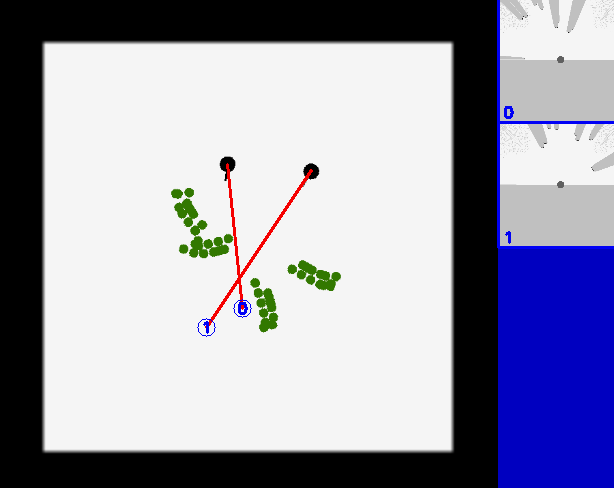}\label{fig:example:e2}}
        \caption{Two scenarios for the training, where the blue digital circles represent the target positions of robots with the corresponding number, red lines specify the straight paths from the starting position to the target for robots, green spots denote the legs' trajectories of pedestrians, black blocks denote static obstacles, and blue boxes on the right illustrate sensor maps of each robot.}
        \label{fig:example}
\end{figure}

As illustrated in Fig.~\ref{fig:example}, we specify two scenarios for the training, i.e., random scenario and circular scenario.
In particular, an environment in random scenario contains two robots, four pedestrians, and four static obstacles, which would randomly choose locations
for obstacles, the starting and target positions of robots and pedestrians.
Then the robots in the environment would be driven by the policy network and pedestrians would be driven either by ORCA or SFM. 
Note that, there are two robots sharing the same navigation policy in the environment and they also need to avoid collisions with each other.
As discussed in \cite{chen2020distributed}, this helps the network to learn the ability of multi-robot obstacle avoidance, which can also help the robot to avoid pedestrians.
Training in environments of random scenario enables the policy network to avoid static and dynamic obstacles, and interact with pedestrians.
An environment in circular scenario contains two robots and four pedestrians, which would randomly place robots and pedestrians on a circle with a random radius.
Training in environments of circular scenario enhances the policy network to interact with pedestrians.

In the training process, we extend PPO to collect experiences from four different environments in parallel at each iteration.
In particular, we construct two environments of random scenario and drive pedestrians in both environments with ORCA and SFM, respectively.
We also construct two environments of circular scenario with pedestrians driven by ORCA and SFM, respectively.
Then we train the network based on experiences generated from these four environments.

\section{Experiments}
\label{exp}

In this section, we evaluate our PPO based crowd navigation approach in both the simulation and the real world. 
We first specify details of our implementation including the hyper-parameters, hardware, and software for the training.
Then, we quantitatively evaluate the performance of our crowd navigation policy in various simulation scenarios and compare it with other approaches. 
We also deploy the trained model to a differential drive robot and test its navigation performance in the real world.
Both qualitative and quantitative experiments show that our approach allows the robot to successfully interact with various pedestrians and is effective with a high success rate. The demonstration video is also available.

\subsection{Reinforcement Learning Setup}

We trained our crowd navigation policy for the differential drive robot following the PPO algorithm with the hyper-parameters listed in Table~\ref{table1}.

\begin{table}[htbp]
        \caption{Hyper-parameters of the training algorithm}\label{table1}
        \centering
        \newcommand{\tabincell}[2]{\begin{tabular}[t]{@{}#1@{}}#2\end{tabular}}
        \begin{tabular}[t]{rl}
                \toprule
                Hyper-parameter                                              & Value                                  \\
                \midrule
                \tabincell{c}{learning rate for policy}                &
                \tabincell{c}{$\textbf{5} \times 10^{-5} $}                                                      \\
                \tabincell{c}{learning rate for value}                 &
                \tabincell{c}{\textbf{$1 \times 10^{-3}$}}                                                      \\
                \tabincell{c}{discount factor ($\gamma$})              & \tabincell{c}{\textbf{$0.99$}}         \\
                \tabincell{c}{replay buffer size } & \tabincell{c}{\textbf{$2048$}}         \\
                \tabincell{c}{image size}                              & \tabincell{c}{\textbf{$48 \times 48$}} \\
                \tabincell{c}{maximum episode length}                      & \tabincell{c}{\textbf{$200$}}          \\
                \tabincell{c}{robot radius }                       & \tabincell{c}{\textbf{$0.17$}}          \\
                \tabincell{c}{maximum pedestrian speed}                & \tabincell{c}{\textbf{$0.5$}}          \\
                \bottomrule
        \end{tabular}
\end{table}

Both the policy network and the value network are implemented in TensorFlow\footnote{\url{https://www.tensorflow.org/}.} and trained with the Adam optimizer~\cite{kingma2014adam}. 
We also want to train network on MindSpore\footnote{\url{https://www.mindspore.cn/}.} , which is a new deep learning computing framework. These problems are left for future work.
The training hardware is a computer with an i9-9900k CPU and an NVIDIA Titan RTX GPU.

We use PPO-PSC to denote our PPO based approach with continuous actions,
PPO-PSD to denote our PPO based approach with discrete actions.
We compare our approaches to ORCA\footnote{With a slight abuse of notation, we use ORCA here to denote the corresponding crowd navigation policy for the robot.} with perfect sensing and PPO-SD, the modification of our approach PPO-PSD by only using the sensor map.

\begin{figure}
        \centering
        \subfigure[Reward curve]{\includegraphics[width = 0.49\linewidth]{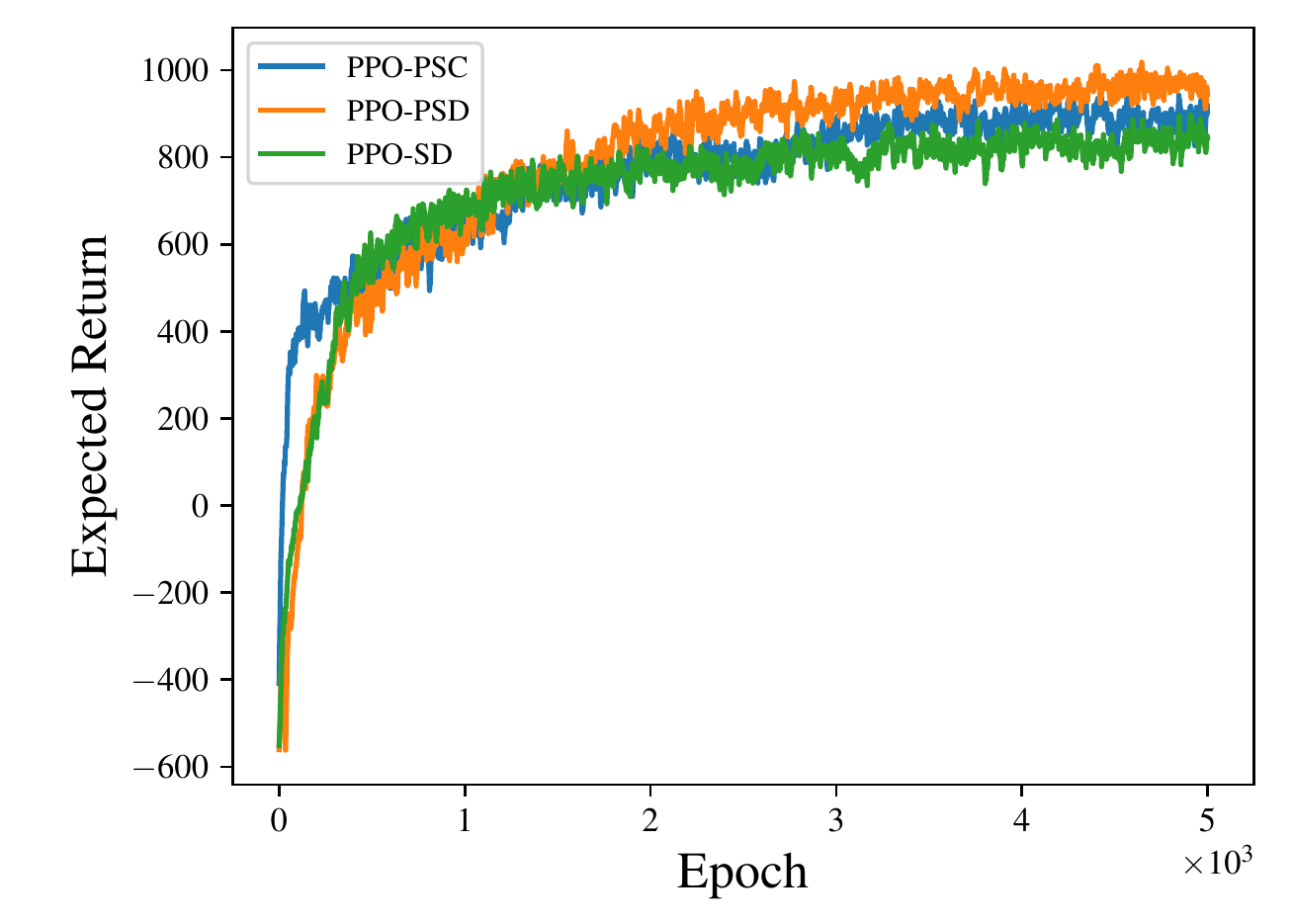}\label{fig:reward:a}}
        \subfigure[Success rate curve]{\includegraphics[width = 0.49\linewidth]{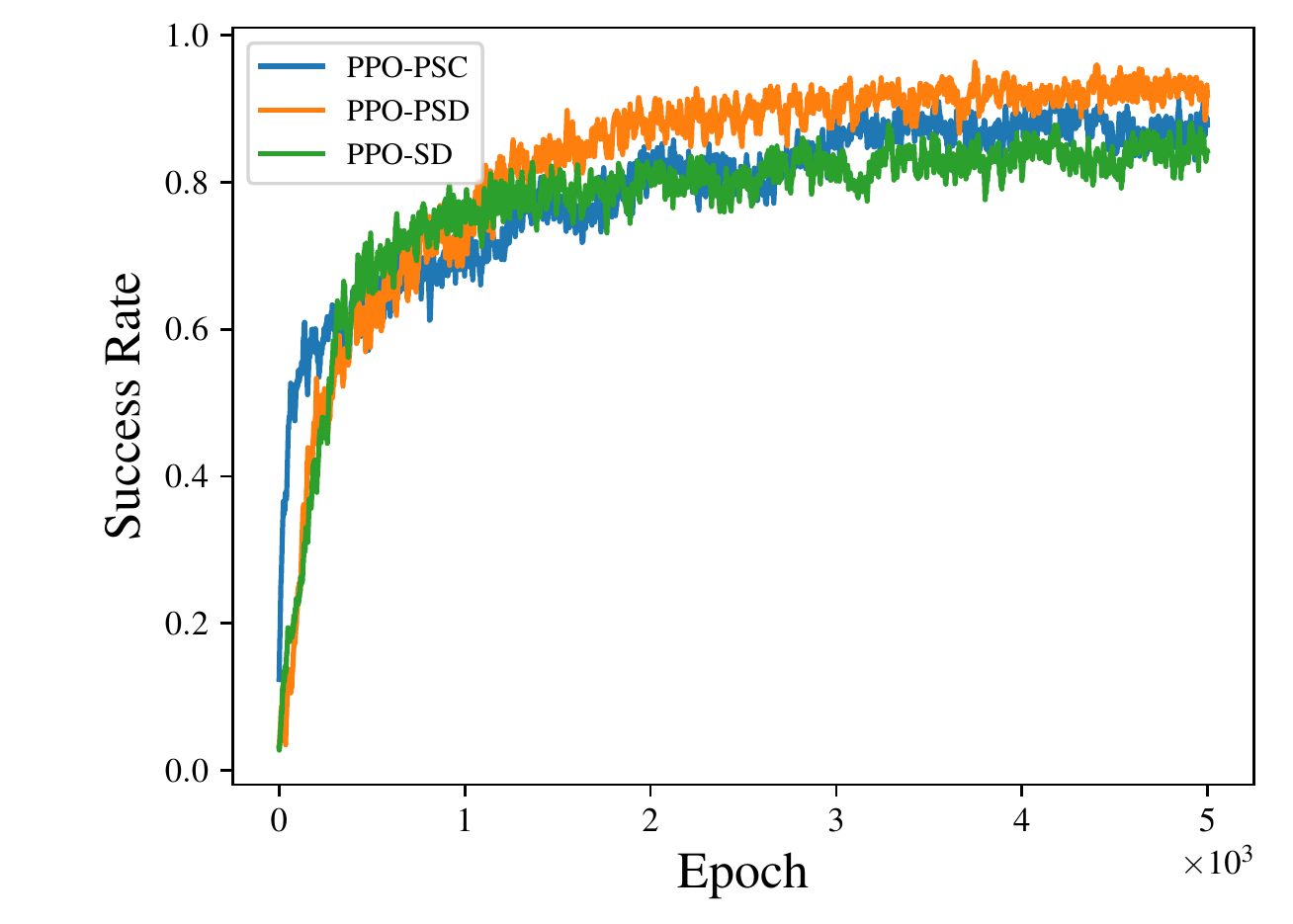}\label{fig:reward:b}}
        \caption{The average reward curve and the success rate curve of approaches.
        }
        \label{fig:reward}
\end{figure}

Fig.~\ref{fig:reward} shows the average reward curve and the success rate curve of the three PPO based approaches, i.e., PPO-PSC, PPO-PSD, and PPO-SD.
It can be seen that the introduction of the pedestrian map can improve the performance of the PPO based crowd navigation.

\subsection{Experiments on Simulation Scenarios}

\subsubsection{Performance metrics}

We introduce three metrics to evaluate the performance of approaches for crowd navigation as the following:
\begin{itemize}
        \item \textit{Success rate ($\bar{\pi }$)}:
              the ratio of the episodes that end with robots reaching their targets without collision.
        \item \textit{Extra time ($\bar{t}$)}:
              the  time required for the robot to successfully reach its target without  collision minus the time for the robot to drive straight to its target  with the maximum speed.
        \item \textit{Average angular velocity change ($\nabla \omega$)}: the average changes of the angular velocity at each step, which reflects the 
        smoothness of the trajectory.
\end{itemize}

\subsubsection{Comparative experiments}

Now we compare the performance of four crowd navigation approaches, i.e., ORCA, PPO-SD, PPO-PSC, and PPO-PSD, in simulation environments of random and circular scenarios. 

Table~\ref{table2} summarizes the average results of four approaches in 500 different environments of each scenario, where pedestrians are driven by ORCA and SFM. Note that the angular velocity change value of ORCA in the table is empty, because the classic ORCA algorithm only outputs the linear velocity in the x and y directions and not the angular velocity.

Note that, ORCA requires every moving agents to cooperate with each other in the environment.
Then there will be collisions between pedestrians in environments, where some pedestrians are driven by ORCA and some are driven by SFM, and these environments are not applicable for testing the performance of crowd navigation approaches. 


\begin{table}[htp]
        \centering
        \caption{Performance of approaches for pedestrians with ORCA and SFM}
        \newcommand{\tabincell}[2]{\begin{tabular}[t]{@{}#1@{}}#2\end{tabular}}
        \begin{tabular}[t]{lrccc}
                \toprule
                Environments                     & Methods                       & $\bar{\pi }$                  & $\bar{t} $            & $\triangledown \omega $                \\

                \midrule
                \multirow{3}*{Random scenario with ORCA}         &
                \tabincell{c}{ORCA} & \tabincell{c}{0.464}          & \tabincell{c}{2.97}  & \tabincell{c}{-} \\       
                \specialrule{0em}{1pt}{1pt}& 
                \tabincell{c}{PPO-SD}       & \tabincell{c}{0.856}   & \tabincell{c}{3.60} & \tabincell{c}{0.63}                  \\
                \specialrule{0em}{1pt}{1pt} & 
                \tabincell{c}{PPO-PSC} & \tabincell{c}{0.862}  & \tabincell{c}{5.64} & \tabincell{c}{0.74}  \\
                \specialrule{0em}{1pt}{1pt} & 
                \tabincell{c}{PPO-PSD} & \tabincell{c}{0.936}  & \tabincell{c}{4.83}  & \tabincell{c}{0.57} \\
                \midrule
                \multirow{3}*{Circular scenario with ORCA}         &
                \tabincell{c}{ORCA} & \tabincell{c}{0.862}    & \tabincell{c}{7.54} & \tabincell{c}{-}\\
                \specialrule{0em}{1pt}{1pt}& 
                \tabincell{c}{PPO-SD}       & \tabincell{c}{0.916}   & \tabincell{c}{4.25}& \tabincell{c}{0.82}                  \\
                \specialrule{0em}{1pt}{1pt} & 
                \tabincell{c}{PPO-PSC} & \tabincell{c}{0.948}  & \tabincell{c}{5.18}& \tabincell{c}{1.02}   \\
                \specialrule{0em}{1pt}{1pt} & 
                \tabincell{c}{PPO-PSD} & \tabincell{c}{0.994}  & \tabincell{c}{4.87} & \tabincell{c}{0.57}  \\
               
                \midrule
                \multirow{3}*{Random scenario with SFM}         &
                \tabincell{c}{ORCA} & \tabincell{c}{0.318}          & \tabincell{c}{1.53}  & \tabincell{c}{-} \\       
                \specialrule{0em}{1pt}{1pt}& 
                \tabincell{c}{PPO-SD}       & \tabincell{c}{0.840}   & \tabincell{c}{3.45} & \tabincell{c}{0.63}                  \\
                \specialrule{0em}{1pt}{1pt} & 
                \tabincell{c}{PPO-PSC} & \tabincell{c}{0.873}  & \tabincell{c}{4.90} & \tabincell{c}{0.74}  \\
                \specialrule{0em}{1pt}{1pt} & 
                \tabincell{c}{PPO-PSD} & \tabincell{c}{0.968}  & \tabincell{c}{4.99}  & \tabincell{c}{0.56} \\
                \midrule
                \multirow{3}*{Circular scenario with SFM}         &
                \tabincell{c}{ORCA} & \tabincell{c}{0.216}    & \tabincell{c}{1.43} & \tabincell{c}{-}\\
                \specialrule{0em}{1pt}{1pt}& 
                \tabincell{c}{PPO-SD}       & \tabincell{c}{0.902}   & \tabincell{c}{4.49}& \tabincell{c}{0.84}                  \\
                \specialrule{0em}{1pt}{1pt} & 
                \tabincell{c}{PPO-PSC} & \tabincell{c}{0.924}  & \tabincell{c}{6.22}& \tabincell{c}{0.97}   \\
                \specialrule{0em}{1pt}{1pt} & 
                \tabincell{c}{PPO-PSD} & \tabincell{c}{0.996}  & \tabincell{c}{5.90} & \tabincell{c}{0.63}  \\
                \bottomrule
                \label{table2}
        \end{tabular}
\end{table}

Table~\ref{table2} shows that, PPO based approaches allow the robot to successfully interact with pedestrians driven either by ORCA or SFM. 
Moreover, the introduction of the pedestrian map can improve the performance. 
PPO-PSD outperforms others in terms of the success rate.

\begin{figure}
        \centering
        \subfigure[Trajectories by ORCA]{\includegraphics[width = 0.46\linewidth]{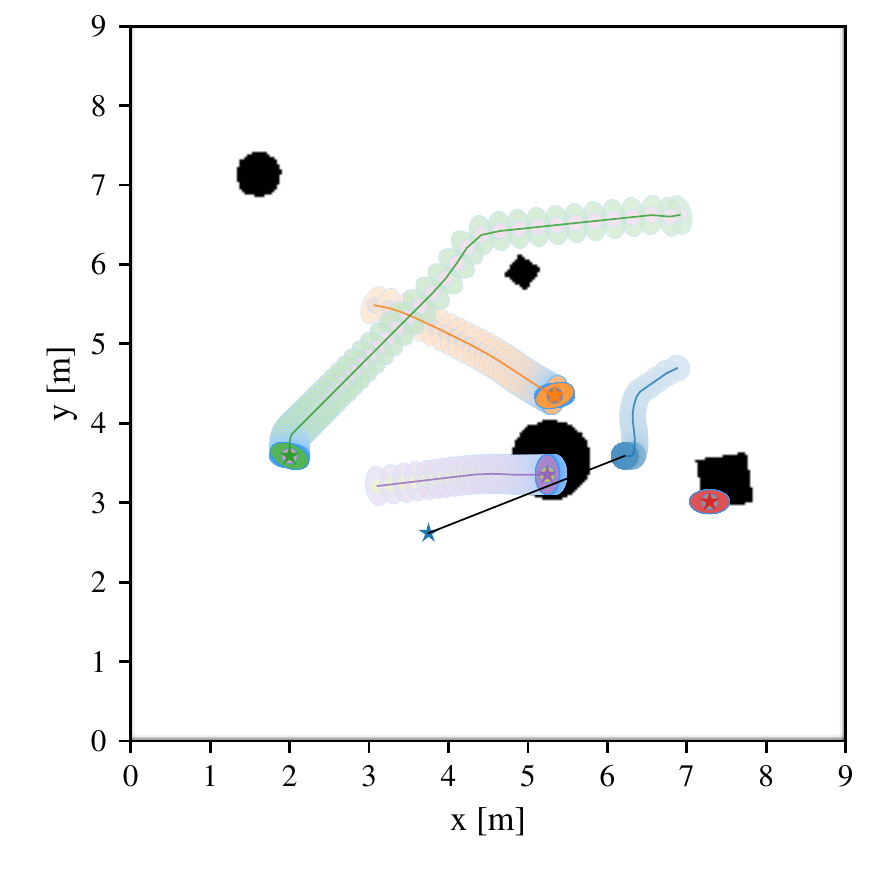}}
        \subfigure[Trajectories by PPO-SD]{\includegraphics[width = 0.46\linewidth]{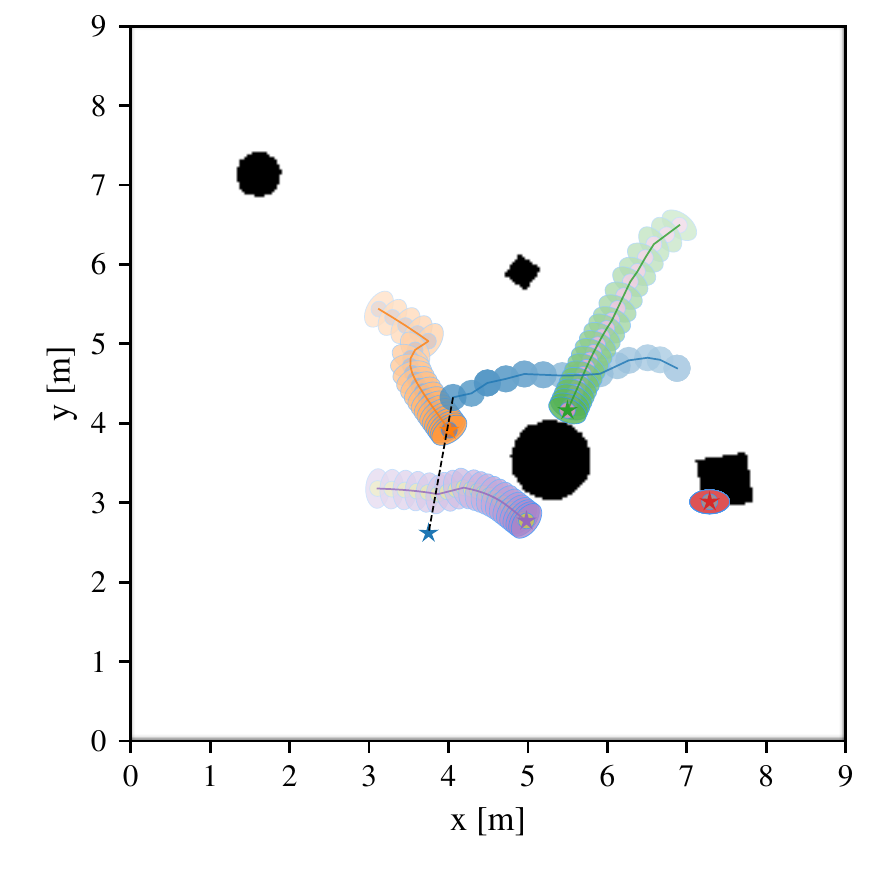}}
        \subfigure[Trajectories by PPO-PSC]{\includegraphics[width = 0.46\linewidth]{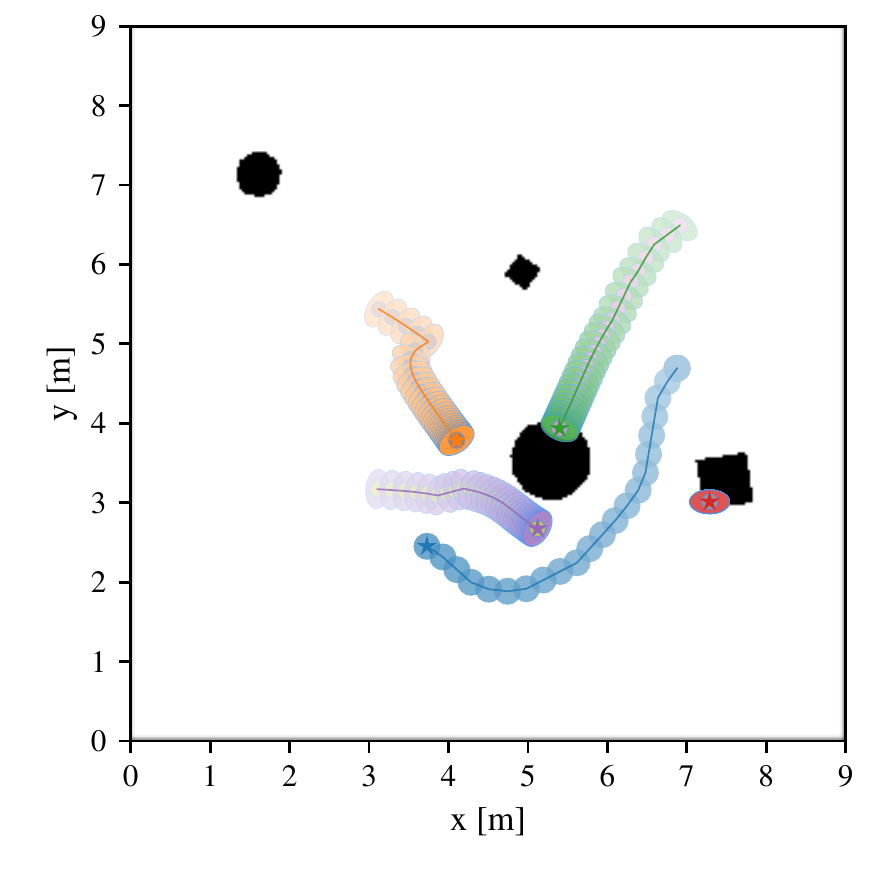}}
        \subfigure[Trajectories by PPO-PSD]{\includegraphics[width = 0.46\linewidth]{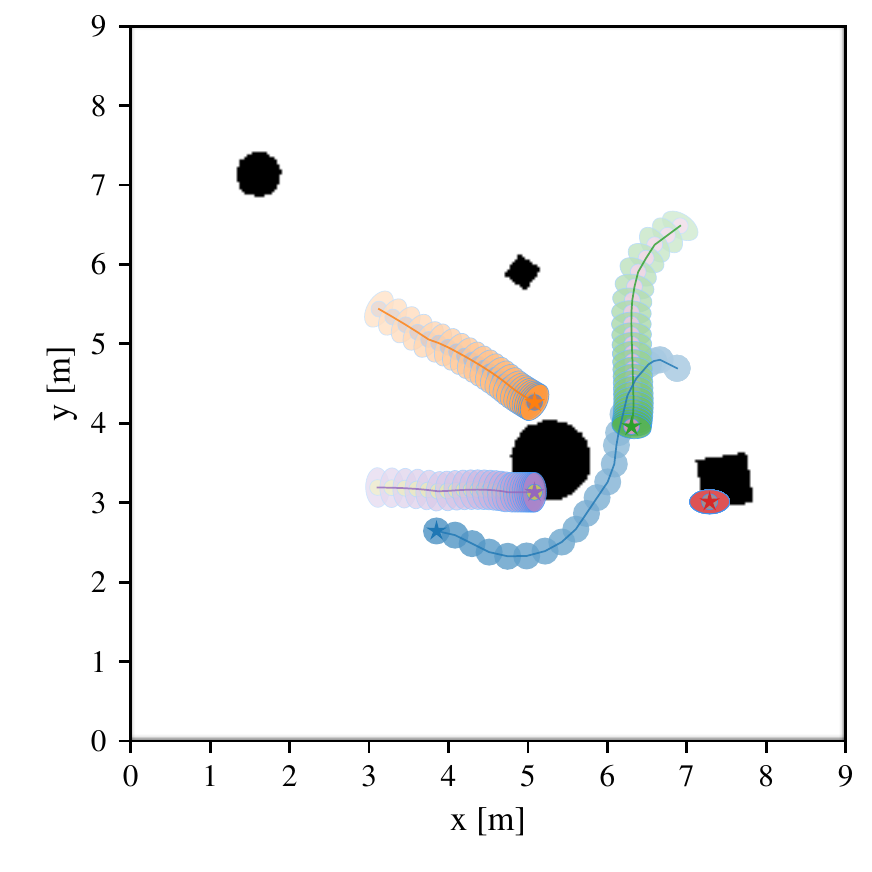}}
        \caption{Robot's trajectories generated by different approaches. The blue spots denote the trajectories of the robot and other spots denote the trajectories of pedestrians.}
        \label{fig:trajec}
\end{figure}

Fig.~\ref{fig:trajec} illustrates robot's trajectories generated by different approaches, where the trajectory generated by ORCA leads the robot to be stuck due to nearby obstacles and pedestrians, the trajectory generated by PPO-SD leads the robot to collide with a pedestrian as the pedestrian map is not considered in it, and both PPO-PSC and PPO-PSD successfully drive the robot to its target. 

The experimental results show that our approach PPO-PSD is effective and outperforms others.

\subsubsection{Pedestrian with multiple strategies}

As specified in Section~\ref{sec:train}, we train the network on both environments with pedestrians driven by ORCA and SFM, respectively.
Here we compare this training process with alternatives.

Most existing crowd navigation approaches assume that pedestrians follow a predefined collision avoidance strategy and test their performance under the same strategy. 
Here we share the configurations for the training environments, except the collision avoidance strategies for pedestrians.
We introduce two variants of our PPO-PSD approach, i.e., PPO-ORCA and PPO-SFM.
In particular, PPO-ORCA is modified from PPO-PSD by only considering pedestrians driven by ORCA, and PPO-SFM is modified from PPO-PSD by only considering SFM.
With a slight abuse of notation, we also use PPO-multi to denote our approach PPO-PSD.

We evaluate the three approaches, i.e., PPO-multi, PPO-ORCA, and PPO-SFM, in environments of five scenarios.
In specific, ORCA-random (resp. SFM-random) denotes the scenario from random scenario by only considering pedestrians driven by ORCA (resp. SFM), ORCA-circular (resp. SFM-circular) denotes the scenario from circular scenario by only considering pedestrians driven by ORCA (resp. SFM), and PPO-circular denotes the scenario that only contains five robots in its environment and randomly places the starting and target positions of the five robots on a circle with a random radius.

\begin{figure}
        \centering
        \subfigure{\includegraphics[width = 0.9\linewidth]{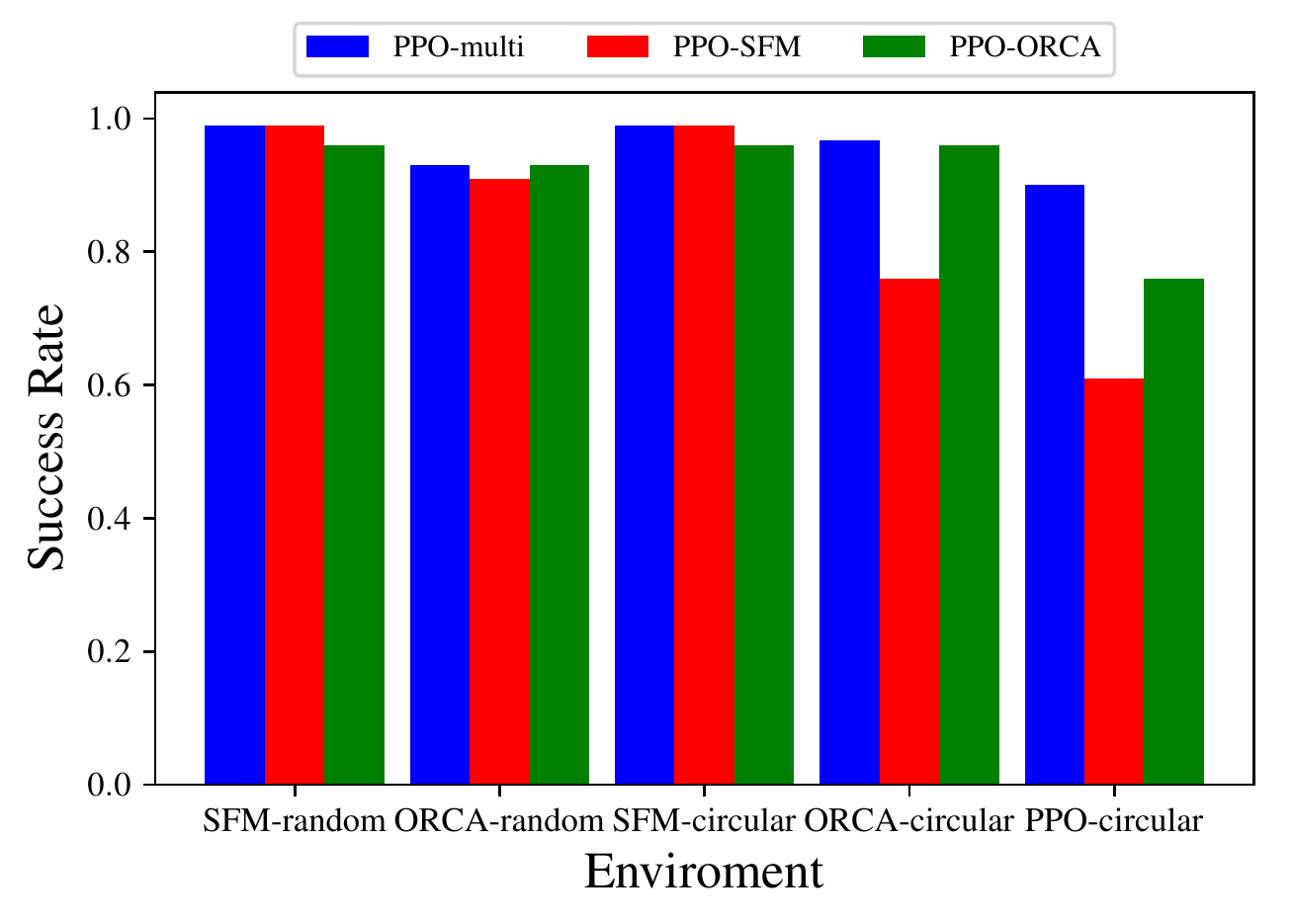}}
        \caption{Success rate of three approaches on five scenarios.
        }
        \label{fig:env}
\end{figure}

The performance of the three approaches on these five scenarios is shown in Fig.~\ref{fig:env}, i.e., the average success rate of each approach in 500 different environments of each scenario.
The results show that PPO-SFM outperforms PPO-ORCA on SFM-random and SFM-circular, PPO-ORCA outperforms PPO-SFM on ORCA-random and ORCA-circular, and both of the approaches do not perform well on PPO-circular.
PPO-multi (PPO-PSD) outperforms all other approaches in environments of all five scenarios, which implies that our training process is effective and allows the robot to successfully interact with pedestrians driven either by ORCA or SFM.

\subsection{Deployment on Actual Robot}

We deploy the trained model by PPO-PSD to an actual differential drive robot to perform crowd navigation in the real world.

\begin{figure}[htp]
        \centering
        \begin{overpic}[ width=0.9\linewidth]{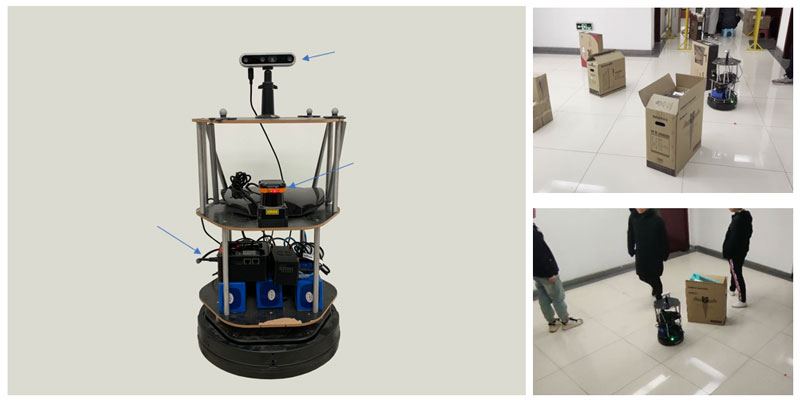}
                \put(45,29.5){\footnotesize laser scanner}
                  \put(43,43){\footnotesize depth camera}
                 \put(2.5,21){\footnotesize Jetson TX2}
                 
        \end{overpic}
        \caption{The robot and its evaluation environments.
        Environments with paper boxes to serve as static obstacles (upper right) and walking pedestrians (bottom right).}
        \label{fig:turtle}
\end{figure}

As shown in Fig.~\ref{fig:turtle}, the robot is based on TurtleBot 2 with Kobuki base and uses a Hokuyo UTM-30LX scanning laser Rangefinder as the 2D laser sensor, a RealSense D455 depth camera as the pedestrian tracking sensor.
The robot equips with two NVIDIA Jetson TX2 for the computation, where one is used to detect and track pedestrians and the other is used to run the trained network.

We use SPENCER people tracking framework~\cite{linder2016multi} in our implementation to track surrounding pedestrians of the robot and generate the pedestrian map for the model.
We also replace the people detection module in SPENCER by YOLOv3~\cite{redmon2018yolov3} for the better detection.
We use the costmap generated by the laser sensor to construct the sensor map. 
Both the sensor map and the pedestrian map have the fixed size $6.0 \times 6.0$m with the resolution 0.125m.
We apply a particular filter based state estimator to generate robot's local targets from its global path in navigation. 
Then we deploy the trained model to be fed with these inputs and let the differential robot execute the outputs, i.e., desired linear and angular velocities, directly.

We evaluate the performance of crowd navigation of the robot in following real-world scenarios.
\begin{itemize}
        \item \textit{Static scenario}: environments that randomly place multiple paper boxes and a suitcase to block the robot from its starting position to its target.
        \item \textit{Dynamic scenario}: environments that the robot needs to pass through a hall while some pedestrians walking around the robot and some paper boxes on its way.
        \item \textit{Corridor scenario}: environments that the robot needs to pass through a corridor while some pedestrians walking towards the robot.
\end{itemize}
Notice that, the corridor in our experiments is quite narrow, then it is challenging for crowd navigation in environments of corridor scenario.

\begin{figure}[htp]
        \centering
        \subfigure[Simple dynamic scenario]  {\includegraphics[width = 0.9\linewidth]{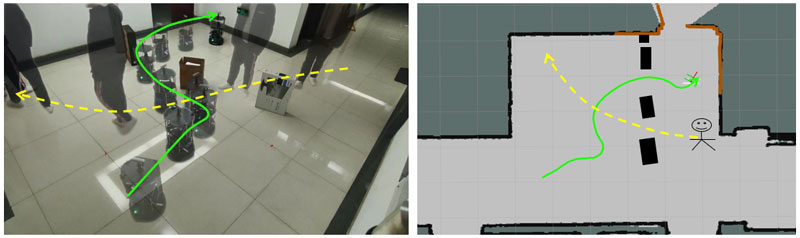}\label{fig:real_tra1:a}}
        \subfigure[Dense dynamic scenario] {\includegraphics[width = 0.9\linewidth]{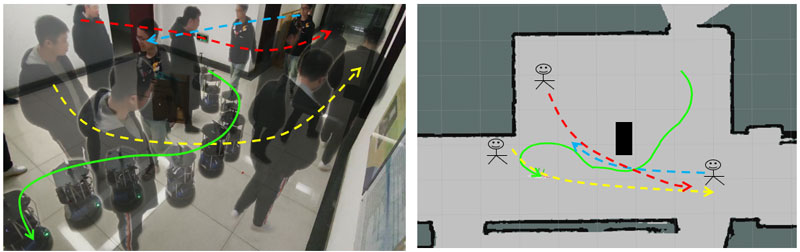}\label{fig:real_tra1:b}}
        \caption{Trajectories of the robot in environments of dynamic scenario. 
        (a) illustrates a simple environment where the global path of the robot is a straight line to the target and the network drives the robot to avoid the obstacles and the walking pedestrian.
        (b) illustrates a dense environment with multiple pedestrians and static obstacles, where the network still needs to drive the robot to avoid pedestrians walking towards it.}
        \label{fig:real_tra1}
\end{figure}

\begin{figure}
        \centering
        \subfigure[Corridor scenario] {\includegraphics[width = 0.9\linewidth]{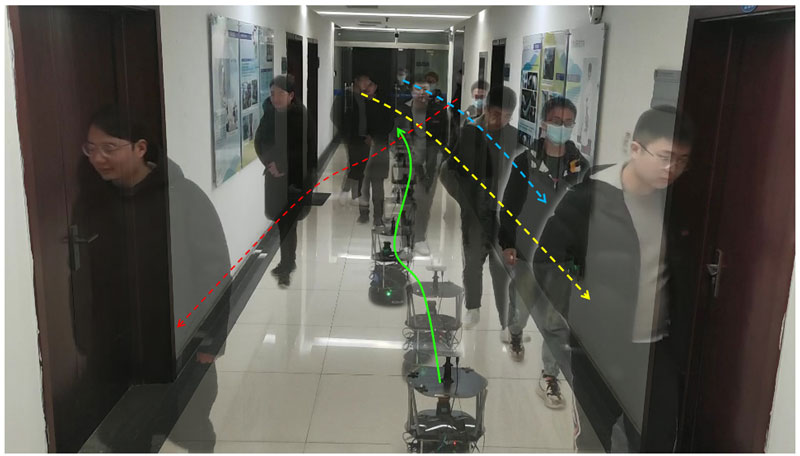}\label{fig:real_tra2:a}}
        \subfigure[Camera view] {\includegraphics[width = 0.9\linewidth]{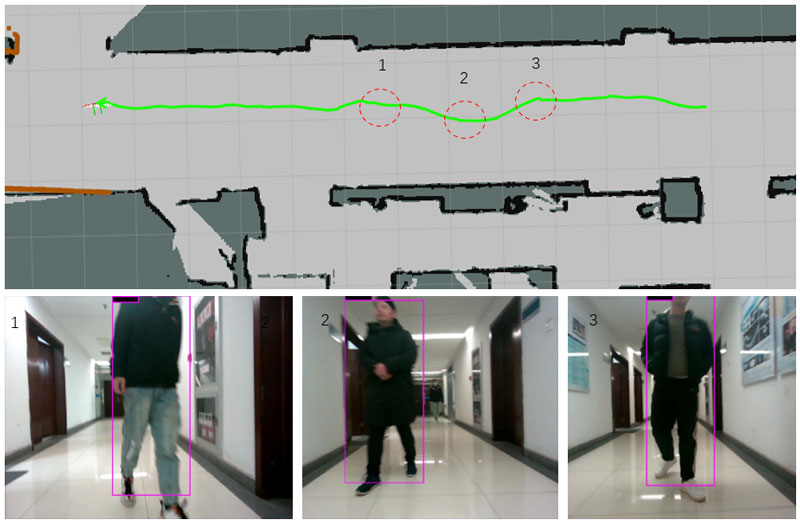}\label{fig:real_tra2:b}}
        \caption{Trajectories of robots in an environment of corridor scenario. 
        (a) illustrates the trajectories of the robot and the surrounding walking pedestrians.
        (b) denotes the map of the environment and corresponding trajectory of the robot (upper image), and illustrates the camera views when the robot is at the corresponding dotted circles in the map (bottom image).}
        \label{fig:real_tra2}
\end{figure}

Our experiments show that the robot can interact with various pedestrians successfully and accomplish navigation tasks safely in most environments of these three scenarios.
Fig.~\ref{fig:real_tra1} and~\ref{fig:real_tra2} illustrate the performance of the robot on the scenarios.
More experiments can be found in our demonstration video.
Note that, in few cases, the robot did not avoid the pedestrian in front of it. 
As discussed in~\cite{fan2020distributed}, a safety strategy can be added to ensure the safety.  


\section{CONCLUSIONS}
\label{conclusion}

In this paper, we argue that pedestrian with multiple different collision avoidance strategies need to be handled by crowd navigation approaches. 
To address this challenge, we propose a PPO based crowd navigation approach that allows pedestrians driven by either ORCA or SFM.
In particular, we apply both the sensor map and the pedestrian map
as the inputs of the neural network. 
We show that distinguishing the information between regular obstacles and pedestrians improves the performance of crowd navigation.
We also train the network in both simulation environments with pedestrians following ORCA and SFM respectively.
We show that this training process allows the robot to successfully interact with various pedestrians.
We also deploy the trained model to an actual robot and evaluate its performance in the real world.
The experimental results show that our approach, PPO-PSD, is effective and outperforms compared methods in terms of the success rate.

Our work makes an improvement towards rapid and successful crowd navigation. Moving forward, we will consider more collision avoidance strategies for strategies and further investigate the generalization of the trained model.

\bibliographystyle{IEEEtran}
\bibliography{IEEEabrv,my}
\end{document}